\documentclass[conference]{IEEEtran}
\IEEEoverridecommandlockouts
\usepackage{caption}
\usepackage{cite}
\usepackage{placeins} 
\usepackage{subcaption}
\usepackage{url}
\usepackage{multirow}
\usepackage{amsmath,amssymb,amsfonts}
\usepackage{algorithmic}
\usepackage{graphicx}
\usepackage{textcomp}
\usepackage{xcolor}
\def\BibTeX{{\rm B\kern-.05em{\sc i\kern-.025em b}\kern-.08em
    T\kern-.1667em\lower.7ex\hbox{E}\kern-.125emX}}
\begin{document}

\title{Physical Evaluation of Naturalistic Adversarial Patches for Camera-Based Traffic-Sign Detection}

\author{
\IEEEauthorblockN{Brianna D'Urso, Tahmid Hasan Sakib, Syed Rafay Hasan, Terry N Guo}
\thanks{

B. D'Urso is with the Department of Computing Sciences, University of Hartford, West Hartford, CT, USA (email: jdurso@hartford.edu).  T.H. Sakib, and S.R. Hasan are with the Department of Electrical and Computer Engineering, Tennessee Technological University, Cookeville, TN, USA (email: tsakib42, shasan\{@tntech.edu\}).  Terry N Guo is with the Center for Manufacturing Research, Tennessee Technological University, Cookeville, TN, USA. (e-mail: nguo@tntech.edu). This research is partially supported by the Tennessee Tech University’s Center for Manufacturing Research, and National Science Foundation Grant (NSF-REU
2349104)
}
}

\maketitle

\begin{abstract}
This paper studies how well Naturalistic Adversarial Patches (NAPs) transfer to a physical traffic sign setting when the detector is trained on customized dataset for autonomous vehicle (AV) environment. We construct a composite dataset, CompGTSRB (which is customized dataset for AV environment), by pasting traffic sign instances from the German Traffic Sign Recognition Benchmark (GTSRB) onto undistorted backgrounds captured from the target platform. CompGTSRB is used to train YOLOv5 model and generate patches using a Generative Adversarial Network (GAN) with latent space optimization, following existing NAP methods. We carried out a series of experiments on our Quanser QCar testbed utilizing the front CSI camera provided in QCar. Across configurations, NAPs reduce the detector’s STOP class confidence. Different configurations includes distance, patch sizes, and patch placement. These results along with a detailed step-by-step methodology indicate the utility of CompGTSRB dataset and the proposed systematic physical protocols for credible patch evaluation. The research further motivate reseaching the defenses that address localized patch corruption in embedded perception pipelines.
\end{abstract}


\section{Introduction} \label{sec:introduction}
Traffic sign perception is a core component of camera based autonomy and driver assistance in autonomous vehicle (AV). It supports rule compliance and route execution, and it often serves as a trigger for downstream control actions. Because these systems operate in open environments, they are exposed to adversarial inputs that can be physically placed in the scene. Adversarial patches are a practical example. A printed pattern can be attached to a target object and can reduce a model’s confidence or change its predicted label without modifying the camera or the model~\cite{b1,b2}.

Prior work shows that patch attacks can transfer beyond synthetic overlays, but physical success is sensitive to viewpoint, distance, lighting, and camera effects such as blur and lens distortion~\cite{b2,b3}. Many evaluations still rely on digitally applied patches, controlled image capture, or datasets whose object presentation differs from what an onboard camera observes during operation~\cite{b1,b2}. 

Recent work has proposed “naturalistic” patches that optimize in the latent space of a pretrained generative model rather than directly over pixels, with the goal of producing patterns that look less like noise and that survive printing and imaging~\cite{b4}. In this setting, a generator such as BigGAN (Generative Adversarial Network) provides a high fidelity image prior~\cite{b5}, and latent direction methods help navigate the generator space during optimization~\cite{b6}. Naturalistic patches have improved physical plausibility in several demonstrations, but there remains a need for evaluations that combine (i) a detector trained on data that matches the target camera, and (ii) a systematic physical test over viewing factors that determine whether the patch is visible and influential.

A central practical issue is dataset mismatch. Public traffic sign datasets such as the German Traffic Sign Recognition
Benchmark (GTSRB) contain many centered, relatively clean sign crops and do not reflect the full range of backgrounds, sign sizes, and off axis views seen by an onboard camera~\cite{b7}. When a detector is trained on such data and then deployed, errors can arise because the operational distribution differs from the training distribution. This is especially relevant for physical patch studies, because the apparent size and placement of both the sign and the patch change strongly with distance and perspective. To reduce this mismatch, our work constructs a composite dataset by pasting traffic sign instances onto backgrounds captured from the target platform, and by applying camera calibration operations so that the composite images preserve lens distortion consistent with the real camera~\cite{b8}.

\begin{figure*}[!htbp]
  \centering
  \includegraphics[width=\textwidth, trim=0.7cm 0.05cm 0.7cm 0.05cm,clip]{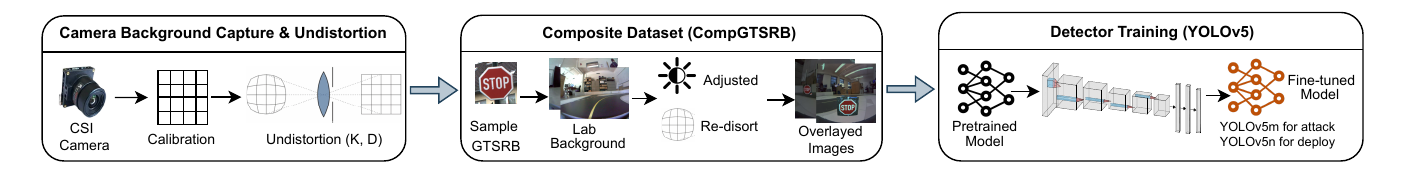}
  \caption{\small Background capture and composite dataset (CompGTSRB) generation, followed by YOLOv5 training for attack and deployment models using the same composite data.}
  \label{fig:pipeline}
\end{figure*}

This paper evaluates naturalistic adversarial patches (NAP) in a complete embedded pipeline. We use the STOP sign class as a representative safety-critical target because detections can directly trigger braking decisions, and missed or delayed detections are easy to interpret at the system level. The same workflow can be applied to other traffic-sign classes, but we focus on STOP sign to keep the evaluation controlled. We construct a camera matched (i.e. images are aligned with the physical characteristics of the specific camera that captured the data) composite dataset derived from GTSRB and platform backgrounds, train a YOLOv5 based detector on this data for embedded deployment~\cite{b9}, generate naturalistic patches following Hu et al.~\cite{b4}, and physically test printed patches on a Quanser QCar using its front CSI camera~\cite{b10}. The study varies distance, patch size, and patch placement on the sign, and it compares learned naturalistic patches against simple occlusion baselines. It is observed that the naturalistic patches can reduce detector confidence in the physical setting, but the effect is moderate and depends on viewing conditions. Simple occluders can be competitive under some distances and placements, which motivates careful baselines and reporting practices when claiming physical robustness or vulnerability.

The contributions of this paper are as follows:
\begin{itemize}
    \item We construct a composite, camera matched traffic sign dataset (called as CompGTSRB) by combining platform captured backgrounds with GTSRB sign instances and camera calibration transforms.
    \item We integrate a naturalistic patch generation pipeline with a YOLOv5 detector trained on this composite data, then evaluate printed patches over different distance, patch size, and patch placement on an embedded platform.
\end{itemize}

\section{Related Works}
Early work on adversarial patches showed that localized patterns can reliably change a model’s prediction under common image transformations, which makes patches a practical physical threat~\cite{b1}. Follow up studies demonstrated printed attacks in real scenes, including road sign scenarios where viewpoint, distance, and lighting can determine whether the attack transfers beyond a digital overlay~\cite{b2}. For object detection, the problem is often harder because the model must both localize and classify the target, and physical success depends on how the patch interacts with the detector’s feature hierarchy~\cite{b3}. Physical detector focused studies report that camera effects and scene geometry can substantially change outcomes, so evaluations should specify the sensing conditions rather than reporting a single aggregate result~\cite{b11}.

Several studies focus specifically on traffic signs and driving relevant perception \cite{b17}. GTSRB remains a common benchmark for traffic sign research, but its image statistics do not fully match onboard camera operation, especially for small and off axis signs in cluttered backgrounds~\cite{b7}. Recent benchmarks and evaluations therefore emphasize realism, including patch studies that model pose and capture variations more directly on sign like targets~\cite{b12}. Other work argues that component level evasion does not always translate to system level impact in autonomy stacks, which motivates reporting conditions that affect visibility and downstream behavior, not only a single confidence value~\cite{b13}. Detector focused driving security studies also incorporate camera transformation models such as scale, blur, and resolution during optimization to improve physical transfer, which further supports the need for test protocols that vary distance and viewing geometry~\cite{b14}.

Naturalistic patch methods aim to improve physical plausibility by optimizing in the latent space of a pretrained generator rather than directly over pixels. Hu et al. propose a naturalistic physical patch pipeline for object detectors that uses a generative prior and shows improved printability compared to unconstrained patterns~\cite{b4}. This approach is consistent with the use of large scale generator priors and latent space navigation methods that enable structured search for effective patterns~\cite{b5,b6}. In parallel, defenses such as PatchGuard study how architectural choices and masking can improve robustness to localized patch corruption, indicating that patch vulnerability remains an active security concern beyond attack construction~\cite{b15}. These works motivate evaluations that combine realistic training data, physical testing across viewing factors, and strong baselines, which is the focus of this paper.

\section*{Methodology} \label{methodology}
\subsection{Workflow Overview}
Our methodology is organized as two linked pipelines. Figure~\ref{fig:pipeline} summarizes the data and detector pipeline, from platform background capture and camera calibration through composite dataset (CompGTSRB) generation and YOLOv5 training for both attack and deployment. Figure~\ref{fig:NAP} summarizes the NAP pipeline, which takes the trained detector and composite images as inputs and optimizes a printable patch in the latent space of a pretrained GAN, following Hu et al.~\cite{b4}. After patch selection, we physically evaluate printed patches on a stop sign using the Quanser QCar front CSI camera by varying distance, patch size, and patch placement, while logging the deployed detector confidence.


\subsection{Platform and Camera Calibration}
All data collection and physical testing use a Quanser QCar equipped with a front CSI camera~\cite{b10}. We estimate the camera intrinsics and distortion parameters using a standard calibration procedure with planar targets~\cite{b8}. These parameters are reused throughout the workflow. Background frames are first undistorted before sign compositing, and the same camera model is applied again to re distort composite images so that the training distribution preserves the lens effects observed by the platform.

\subsection{Composite Dataset Construction (CompGTSRB)} 

\begin{figure}[!htbp]
\centering
\includegraphics[width=\linewidth]{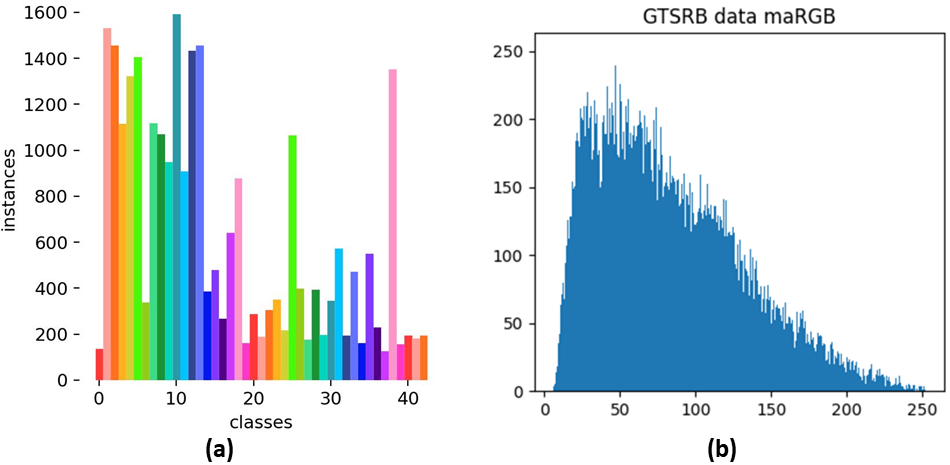}
\caption{\small GTSRB dataset statistics. (a) Class instance counts indicate imbalance. (b) maRGB distribution shows illumination skew that motivates brightness matching in CompGTSRB.}
\label{fig:gtsrb_stats}
\end{figure}

We build CompGTSRB to reduce the gap between public traffic sign data and the platform camera view. GTSRB exhibits class imbalance and illumination skew. Figure~\ref{fig:gtsrb_stats}a  shows the uneven class counts, and figure~\ref{fig:gtsrb_stats}b shows the skewed maRGB distribution. Such skew can bias the detector toward a narrow exposure range, so performance drops when the platform camera produces different brightness due to lighting and auto exposure. These properties can hurt generalization, so we address both effects during compositing.


For each sample, we undistort a QCar background frame so geometric operations are applied consistently, then paste a GTSRB sign with randomized position and scale to reduce artifacts and expose off axis, small-target views typical of onboard sensing. We re-distort the composite using the calibrated camera model and update the bounding box after mapping to keep labels consistent with the projected sign. To reduce illumination shift between sign crops and platform backgrounds, we compute the mean average RGB (maRGB) of the sign and choose a background with the closest maRGB. When reusing samples to address class imbalance, we rescale sign brightness by the background to sign maRGB ratio, clamp values to [0,255], and filter overly dark composites to keep the distribution closer to the target camera range.

\subsection{Detector Training for Attack and Deployment}
We train YOLOv5 detectors on CompGTSRB so that patch generation and deployment share the same data distribution~\cite{b9}. We use two model sizes for different roles. YOLOv5m is used during patch optimization to provide stable gradients, while YOLOv5n is used for embedded deployment on the QCar due to its lower compute cost. Both models are trained using the same labels and image generation process so that performance differences reflect model capacity rather than dataset mismatch.

\subsection{Naturalistic Adversarial Patch Generation}
\begin{figure}[!htbp]
  \centering
  \includegraphics[width=\columnwidth, trim=0.5cm 0.5cm 0.5cm 0.5cm,clip]{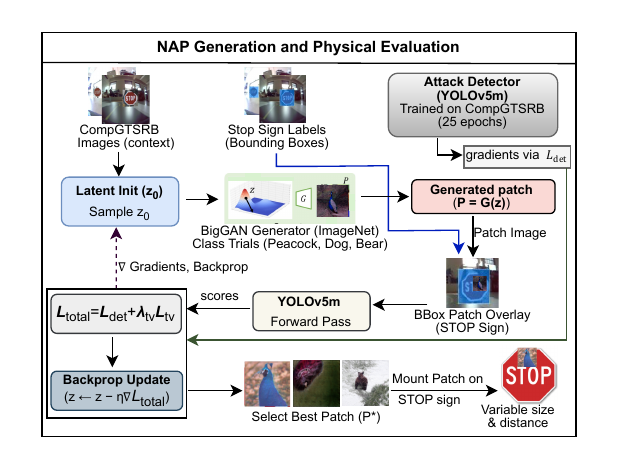}
  \caption{\small NAP generation following Hu et al.~\cite{b4}. A GAN latent vector is optimized to produce a printable patch that lowers detector confidence.}
  \label{fig:NAP}
\end{figure}

We generate naturalistic adversarial patches using the method of Hu et al.~\cite{b4}, which optimizes a printable patch through a pretrained GAN prior rather than directly optimizing pixels. Figure~\ref{fig:NAP} shows the full loop. The inputs are CompGTSRB images and the corresponding stop sign bounding box labels, along with an attack detector. We use a YOLOv5m model trained on CompGTSRB as the attack detector because it provides stable gradients during optimization.

Patch synthesis starts by initializing a latent vector \texttt{z\textsubscript{0}}. This latent is passed through the BigGAN generator \texttt{G} to produce a candidate patch image \texttt{P=G(z)}. We evaluate multiple ImageNet class initializations for the generator, and we report results for three representative trials (Peacock, Dog, Bear). The generated patch is then overlaid onto the stop sign region in the training images using the bounding box location. The patched image is forwarded through YOLOv5m, and we compute a detector loss $L_{\text{det}}$ that penalizes high STOP class confidence. In Figure~\ref{fig:NAP}, we optimize the latent code $z$ by minimizing $L_{\text{total}} = L_{\text{det}} + \lambda_{\text{tv}} L_{\text{tv}}$~\cite{b4}. Here, $L_{\text{tv}}$ is a total-variation regularizer that encourages smooth, printable patches, and $\lambda_{\text{tv}}$ controls its weight. We update $z$ with a gradient step $z \leftarrow z - \eta \nabla_{z} L_{\text{total}}$, where $\eta$ is the step size. Gradients are backpropagated through the detector and into the latent vector, while keeping the detector weights fixed, and we update \texttt{z} by gradient descent. This process is repeated for several iterations, producing a sequence of candidate patches.

\begin{table*}[!htbp]
\caption{\small STOP-class confidence for clean signs ($C$) and change under printed patches ($\Delta C=C_{\text{patch}}-C_{\text{clean}}$), averaged over three placements, stratified by patch size and distance $d$. The final block reports the mean over sizes.}
\label{tab:results_by_size_distance}
\centering
\small
\renewcommand{\arraystretch}{1.05}
\begin{tabular}{|c|c|c|c|c|c|c|c|}
\hline
\textbf{Size} & \textbf{Distance $d$ (m)} & \textbf{Clean $C$} & \textbf{White $\Delta C$} & \textbf{Black $\Delta C$} & \textbf{NAP-Peacock $\Delta C$} & \textbf{NAP-Dog $\Delta C$} & \textbf{NAP-Bear $\Delta C$} \\
\hline

\multirow{5}{*}{Small}
& 0.30 & 0.7788 & +0.022 & +0.001 & -0.178 & -0.128 & -0.223 \\
\cline{2-8}
& 0.38 & 0.7105 & +0.021 & -0.005 & +0.006 & -0.036 & -0.008 \\
\cline{2-8}
& 0.45 & 0.8506 & -0.005 & -0.018 & +0.005 & -0.004 & +0.006 \\
\cline{2-8}
& 0.60 & 0.7862 & +0.004 & +0.005 & +0.014 & +0.015 & +0.017 \\
\cline{2-8}
& 0.90 & 0.8993 & -0.028 & -0.009 & -0.014 & -0.007 & -0.020 \\
\hline

\multirow{5}{*}{Medium}
& 0.30 & 0.7788 & -0.197 & -0.199 & -0.279 & -0.288 & -0.342 \\
\cline{2-8}
& 0.38 & 0.7105 & -0.044 & -0.104 & -0.119 & -0.191 & -0.173 \\
\cline{2-8}
& 0.45 & 0.8506 & -0.030 & -0.030 & -0.010 & -0.030 & -0.051 \\
\cline{2-8}
& 0.60 & 0.7862 & -0.021 & -0.074 & +0.012 & +0.016 & -0.022 \\
\cline{2-8}
& 0.90 & 0.8993 & -0.031 & -0.012 & -0.011 & -0.011 & -0.009 \\
\hline

\multirow{5}{*}{Large}
& 0.30 & 0.7788 & -0.270 & -0.232 & -0.359 & -0.323 & -0.358 \\
\cline{2-8}
& 0.38 & 0.7105 & -0.243 & -0.264 & -0.293 & -0.307 & -0.301 \\
\cline{2-8}
& 0.45 & 0.8506 & -0.132 & -0.079 & -0.035 & -0.147 & -0.142 \\
\cline{2-8}
& 0.60 & 0.7862 & -0.060 & -0.146 & -0.092 & -0.041 & -0.106 \\
\cline{2-8}
& 0.90 & 0.8993 & -0.088 & -0.019 & -0.013 & -0.010 & -0.004 \\
\hline


\end{tabular}
\end{table*}

\begin{figure}[!htbp]
  \centering
  \includegraphics[width=\columnwidth]{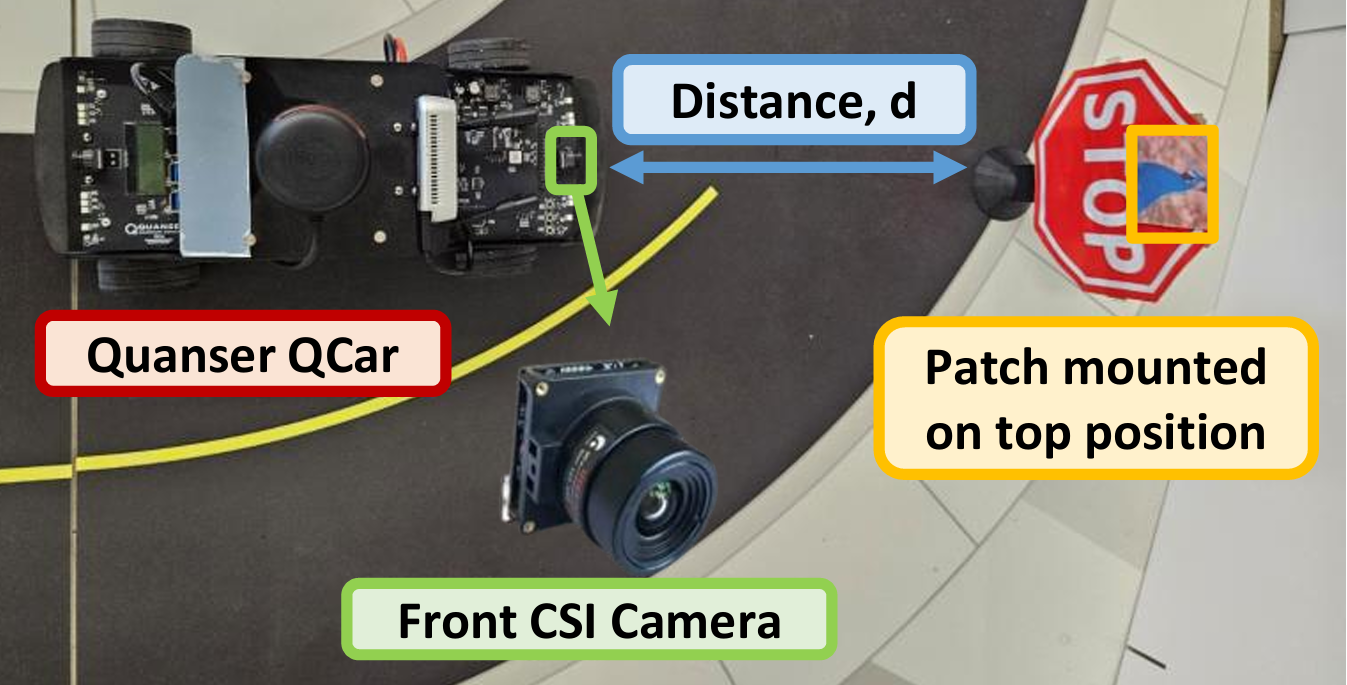}
  \caption{\small Physical setup on the Quanser QCar. Front CSI camera observes a printed stop sign with a mounted patch. Distance \texttt{d} is varied; YOLOv5n runs onboard..}
  \label{fig:setup}
\end{figure}

We select the best patch \texttt{$P^*$} based on the lowest STOP confidence achieved on the optimization set. The selected patch is printed and mounted on a physical stop sign for the evaluation protocol in the next subsection, where we vary patch size and viewing distance during QCar camera captures (see figure~\ref{fig:setup}).

\subsection{Physical Evaluation Protocol}
We evaluate printed patches on a physical stop sign using the Quanser QCar front CSI camera~\cite{b10}. The QCar runs on an NVIDIA Jetson TX2, and YOLOv5n inference is executed onboard in real time. Figure~\ref{fig:setup} shows the top view of the QCar, the camera, the mounted patch, and the definition of distance \texttt{d}. For each trial, a patch is mounted on the sign at a selected placement, and the sign is positioned at a measured distance \texttt{d} from the camera. We vary distance, patch size, and patch placement across trials.


For each configuration, the detector runs for a fixed 15\,s window while the QCar vehicle and STOP sign remain stationary, and we record the per frame STOP confidence output by YOLOv5n. We sweep four factors; patch type (white, black, and three NAP variants), patch size (small, medium, large), patch placement on the sign face (three locations), and camera to sign distance \texttt{d} (five distances). Each entry in Table \ref{tab:results_by_size_distance} reports the mean confidence over the 15 s window, averaged over the three placements for that size and distance.


\section{Experimental Results and Discussion} \label{results}
We evaluate printed patches on a physical stop sign using the Quanser QCar front CSI camera, with YOLOv5n running onboard. Figure~\ref{fig:setup} shows the setup and the camera-to-sign distance (\texttt{d}). For each configuration, we run inference for 15\,s, log the STOP-class confidence, and report the mean confidence (\texttt{C}). Table~\ref{tab:results_by_size_distance} reports ($\Delta \texttt{C} = \texttt{C}_{\text{patch}} - \texttt{C}_{\text{clean}}$) at the same distance, stratified by patch size and averaged over three placements. Negative ($\Delta \texttt{C}$) indicates that the patch reduces detector confidence.

\begin{figure}[!htbp]
\centering
\includegraphics[width=\columnwidth]{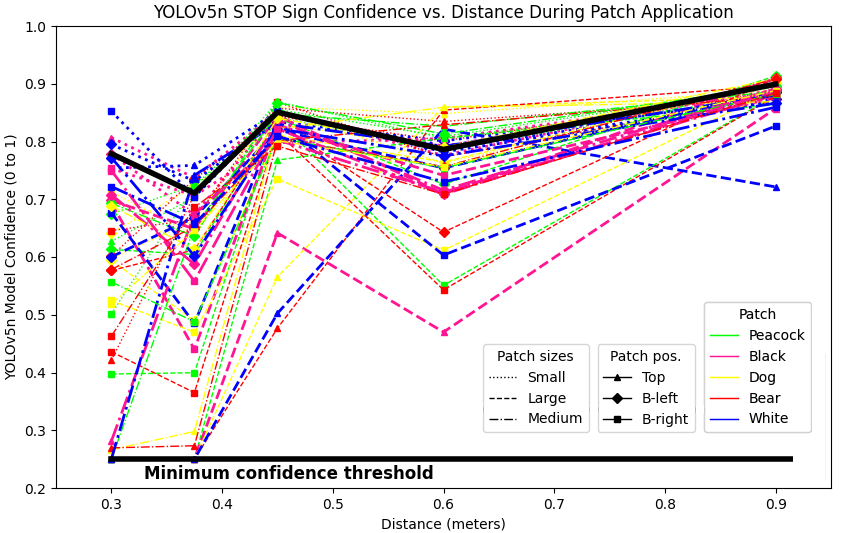}
\caption{Mean YOLOv5n STOP confidence versus distance for clean and patched signs. Results include patch type, size, and placement combinations. Each point is a 15 s mean.}
\label{fig:distance_sweep}
\end{figure}

Figure~\ref{fig:distance_sweep} complements Table~\ref{tab:results_by_size_distance} by showing mean STOP confidence versus distance across all patch type, size, and placement configurations. Separation between clean and patched traces is largest at close range and decreases as distance increases, consistent with reduced patch influence when the sign occupies fewer pixels. The spread at a fixed distance reflects sensitivity to patch size and placement, which motivates the size stratified reporting in Table \ref{tab:results_by_size_distance}. Overall, Figure~\ref{fig:distance_sweep} reinforces that physical patch impact is condition dependent and most observable at short distances.

Table~\ref{tab:results_by_size_distance} shows that patch impact is strongest at close range and weakens as distance increases. Patch size is defined by pixel footprint in the captured frame. At $d = 0.30\,\mathrm{m}$, Large covers $\approx 14.2\%$ of image pixels, while Medium and Small cover $\approx 10.7\%$ and $\approx 6.3\%$, respectively. At $\texttt{d}=0.30\,\texttt{m}$, the large naturalistic patches produce the largest confidence reductions, with NAP values around $-0.32$ to $-0.36$, and the medium patches also reducing confidence substantially. At $\texttt{d}=0.38\,\texttt{m}$, large patches remain effective, while small patches show near-zero changes for several conditions. From $\texttt{d}=0.45\,\texttt{m}$ onward, most effects become modest, and at $\texttt{d}=0.90\,\texttt{m}$ the changes are small across all patch types. This distance trend is consistent with physical constraints reported in prior works, where viewpoint and target scale limit how much a printed patch influences detector features at long range~\cite{b2,b11,b14}.

The table also clarifies a size effect that is masked when averaging over patch sizes. At $\texttt{d}=0.30\,\texttt{m}$, larger patches consistently yield larger-magnitude reductions than medium and small patches for the NAP variants. As distance increases, this advantage shrinks, and several small or medium entries approach zero or become slightly positive. This pattern matches the imaging geometry. A larger patch covers more of the sign and occupies more pixels in the camera frame, so it is more likely to perturb the detector at close range. At longer range, all patches occupy fewer pixels and their influence decreases.

Occlusion baselines remain important for interpreting these results. White and black squares sometimes match or exceed the NAP variants, especially for large patches at mid distances. For example, at $\texttt{d}=0.60\,\texttt{m}$ the large black occluder produces a larger reduction than the large NAP variants in Table~\ref{tab:results_by_size_distance}. This supports two implications. First, the naturalistic optimization produces physically printable patches that can reduce confidence, but the gains over simple occlusion are not uniform. Second, reporting occlusion baselines is necessary when evaluating physical patch attacks, because a measured confidence drop may reflect sign occlusion rather than attack structure.




\section*{Conclusion}
This paper examined whether naturalistic adversarial patches remain effective for traffic sign detection when the full training and deployment pipeline is aligned with the target camera and tested physically on an embedded platform. We addressed dataset mismatch by constructing CompGTSRB, which composites traffic sign instances onto platform captured backgrounds and applies calibration based undistortion and re-distortion so that training images preserve the geometric and illumination properties of the onboard camera. 
The results show that printed patches can reduce STOP class confidence, that can significantly affect the AV's navigation. However, configuration of the experiment lead to different results, e.g. confidence drops are most evident at close range and for larger patches, while effects diminish at longer distances where the patch occupies fewer pixels, and simple occlusion baselines can be competitive under some configurations. Overall, the study supports the need for camera matched data generation and systematic physical protocols when assessing patch robustness on embedded detectors, and it indicates that measured vulnerability should be reported together with viewing conditions and strong baselines. Future work will move beyond component level confidence changes by embedding the attack into a closed loop navigation routine where stop sign detections trigger braking, and by quantifying system level outcomes such as detection distance shifts and stopping error. We will also evaluate patch agnostic mitigation strategies that detect or suppress localized corrupt regions without requiring prior knowledge of the specific patch, alongside improvements to training time robustness through adversarial data augmentation and broader testing across lighting, viewpoint, and additional sign classes. We will
also study patch agnostic defenses that localize and suppress patch regions without patch specific training, including PatchGuard~\cite{b15} and PAD~\cite{b16}.



\begin{thebibliography}{00}
\bibitem{b1} T. Brown, D. Mané, A. Roy, M. Abadi, and J. Gilmer, “Adversarial patch,” arXiv preprint arXiv:1712.09665, 2017.
\bibitem{b2} K. Eykholt et al., “Robust physical-world attacks on deep learning visual classification,” in Proc. IEEE Conf. Computer Vision and Pattern Recognition (CVPR), 2018, pp. 1625–1634.
\bibitem{b3} S. Thys, W. Van Ranst, and T. Goedemé, “Fooling automated surveillance cameras: Adversarial patches to attack person detection,” in Proc. IEEE/CVF Conf. Computer Vision and Pattern Recognition Workshops (CVPRW), 2019.
\bibitem{b4} Y.-C.-T. Hu, C.-H. Hsieh, C.-T. Lin, and W.-H. Cheng, “Naturalistic physical adversarial patch for object detectors,” in Proc. IEEE/CVF Int. Conf. Computer Vision (ICCV), 2021, pp. 7828–7837.
\bibitem{b5} A. Brock, J. Donahue, and K. Simonyan, “Large scale GAN training for high fidelity natural image synthesis,” in Proc. Int. Conf. Learning Representations (ICLR), 2019.
\bibitem{b6} A. Voynov and A. Babenko, “Unsupervised discovery of interpretable directions in the GAN latent space,” in Proc. Int. Conf. Machine Learning (ICML), 2020, pp. 9786–9796.
\bibitem{b7} J. Stallkamp, M. Schlipsing, J. Salmen, and C. Igel, “The German Traffic Sign Recognition Benchmark: A multi-class classification competition,” in Proc. Int. Joint Conf. Neural Networks (IJCNN), 2011, pp. 1453–1460.
\bibitem{b8} Z. Zhang, “A flexible new technique for camera calibration,” IEEE Trans. Pattern Anal. Mach. Intell., vol. 22, no. 11, pp. 1330–1334, 2000.
\bibitem{b9} G. Jocher et al., “YOLOv5,” Ultralytics, 2020. [Online]. Available: https://github.com/ultralytics/yolov5.
\bibitem{b10} Quanser Inc., “QCar,” 2025. [Online]. Available: https://www.quanser.com/products/qcar/.
\bibitem{b11} K. Eykholt, I. Evtimov, E. Fernandes, B. Li, A. Rahmati, F. Tramer, A. Prakash, T. Kohno, and D. Song, “Physical Adversarial Examples for Object Detectors,” in Proc. USENIX Workshop on Offensive Technologies (WOOT), 2018.
\bibitem{b12} N. Hingun, C. Sitawarin, J. Li, and D. Wagner, “REAP: A Large-Scale Realistic Adversarial Patch Benchmark,” in Proc. IEEE/CVF Int. Conf. on Computer Vision (ICCV), 2023. 
\bibitem{b13} N. Wang, Y. Luo, T. Sato, K. Xu, and Q. A. Chen, “Does Physical Adversarial Example Really Matter to Autonomous Driving? Towards System-Level Effect of Adversarial Object Evasion Attack,” in Proc. IEEE/CVF Int. Conf. on Computer Vision (ICCV), 2023. 
\bibitem{b14} W. Jia, Z. Lu, H. Zhang, Z. Liu, J. Wang, and G. Qu, “Fooling the Eyes of Autonomous Vehicles: Robust Physical Adversarial Examples Against Traffic Sign Recognition Systems,” in Proc. Network and Distributed System Security Symposium (NDSS), 2022. 
\bibitem{b15} C. Xiang, A. N. Bhagoji, V. Sehwag, and P. Mittal, “PatchGuard: A Provably Robust Defense against Adversarial Patches via Small Receptive Fields and Masking,” in Proc. USENIX Security Symposium, 2021.
\bibitem{b16} L. Jing, R. Wang, W. Ren, X. Dong, and C. Zou, “PAD: Patch-Agnostic Defense against Adversarial Patch Attacks,” in Proc. IEEE/CVF Conf. Computer Vision and Pattern Recognition (CVPR), 2024, pp. 24472–24481.
\bibitem{b17} A. Solanki, W. Al Amiri, M. Mahmoud, B. Swieder, S. R. Hasan and T. N. Guo, `` Survey of Navigational Perception Sensors’ Security in Autonomous Vehicles,'' IEEE Access, 2025

\end{thebibliography}
\end{document}